\documentclass[10pt,journal,compsoc]{IEEEtran}
\ifCLASSOPTIONcompsoc
  \usepackage[nocompress]{cite}
\else
  \usepackage{cite}
\fi

\usepackage{mathtools}
\usepackage{hyperref}
\usepackage[noabbrev]{cleveref}
\ifCLASSINFOpdf
\else
\fi
\usepackage{amsmath}
\usepackage{amsfonts}
\begin{document}
\title{Domino Denoise: An Accurate Blind Zero-Shot Denoiser using Domino Tilings}

\author{
Jason~Lequyer,
Wen-Hsin~Hsu,
Reuben~Philip,
Anna~Christina~Erpf
and Laurence~Pelletier% <-this % stops a space
\IEEEcompsocitemizethanks{\IEEEcompsocthanksitem All authors are affiliated with the Lunenfeld-Tanenbaum Research Institute, Toronto,
ON.\protect\\

\IEEEcompsocthanksitem J. Lequyer, R. Philip and L. Pelletier are with the Department
of Molecular Genetics, University of Toronto, Toronto,
ON.}% <-this % stops a space
\thanks{Manuscript received XXXX; revised XXXX.}}


\markboth{Journal of \LaTeX\ Class Files,~Vol.~14, No.~8, August~2015}%
{Shell \MakeLowercase{\textit{et al.}}: Bare Advanced Demo of IEEEtran.cls for IEEE Computer Society Journals}



\IEEEtitleabstractindextext{%
\begin{abstract}
Because noise can interfere with downstream analysis, image denoising has come to occupy an important place in the image processing toolbox. The most accurate state-of-the-art denoisers typically train on a representative
dataset. But gathering a training set is not always feasible, so interest has grown in blind zero-shot denoisers that train only on the image they are denoising.
The most accurate blind-zero shot methods are  blind-spot networks, which mask pixels and attempt to infer them from their surroundings. Other methods exist where all neurons participate in forward inference, however they are not as accurate and are susceptible to overfitting. Here we present a hybrid approach. We first introduce a semi blind-spot network where the network can see only a small percentage of inputs during gradient update. We then resolve overfitting by introducing a validation scheme where we split pixels into two groups and fill in pixel gaps using domino tilings. Our method achieves an average PSNR increase of $0.28$ and a three fold increase in speed over the current gold standard blind zero-shot denoiser Self2Self on synthetic Gaussian noise. We demonstrate the broader applicability of Pixel Domino Tiling by inserting it into a preciously published method.
\end{abstract}

}

\maketitle

% is not selected <OR> if conference mode is selected - because compsoc
% conference papers position the abstract like regular (non-compsoc)
\IEEEdisplaynontitleabstractindextext

% \fi
\IEEEpeerreviewmaketitle

\ifCLASSOPTIONcompsoc
\IEEEraisesectionheading{\section{Introduction}\label{sec:introduction}}
\else
\section{Introduction}
\label{sec:introduction}
\fi

\IEEEPARstart{I}{mage} noise is the unavoidable random fluctuation of colour and grayscale intensity values that is present to some degree in virtually all forms of image acquisition. Noise is a nuisance, not only does it make visual inspection more difficult, but it can also interfere with downstream automated analysis such as segmentation \cite{DNSEG,N2F}. It is therefore desirable to reduce the amount of noise in acquired images.

\begin{figure*}
  \centering
  \includegraphics[width=0.9\linewidth]{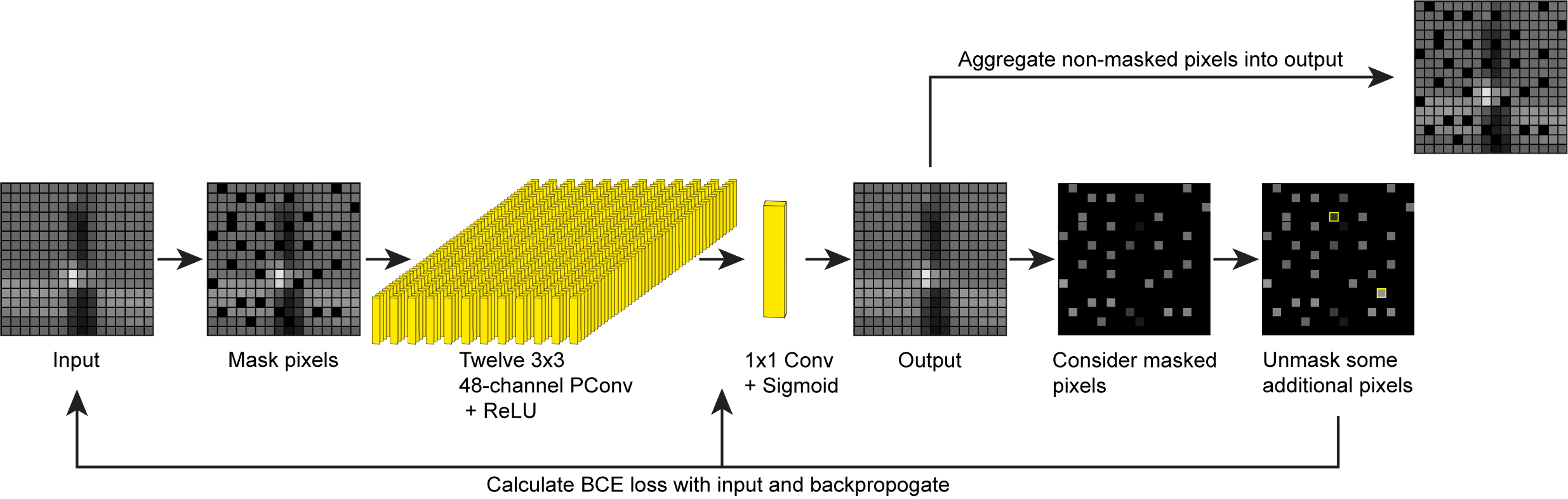}
  \caption{\textbf{Our modified blind-spot network.} At each iteration, $20$ \% of the pixels are randomly masked and sent into our neural network. Network loss is calculated on the original $20$ \% plus an additional $0.1$ \% of the pixels that were not masked. Output pixels are aggregated into pixel-specific lists at each iteration, only unmasked pixels are included. The final output is generated by averaging each pixel's list.}
  \label{fig:1}
\end{figure*}

The simplest way to reduce noise is to minimize its underlying causes, however in many applications this is either not possible or not desirable. For example, in cell microscopy, trade-offs between imaging speed, throughput and sample viability, typically defer the task of noise reduction until post acquisition. Situations like this have given rise to the field of image denoising which is concerned with removing noise from images after they've already been captured and converted into digital formats.

Typically, denoisers work by exploiting the fact that random noise lacks the structure and pattern of clear visual signals, and so the task is to manipulate the image in ways that are biased towards preserving these structures at the expense of noise. While these manipulations can be computed explicitly, the most accurate denoisers to-date learn them using AI \cite{review2}.

The simplest way to do this is to train a neural network to map noisy instances of images to their clean counterparts, as in DnCNN \cite{DnCNN}. However, representative clean/noisy image pairs aren't always available, particularly in microscopy where it is difficult to reduce the noise without also changing the signal \cite{review}. For this reason, more flexible alternatives were developed, such as methods that only require noisy/noisy image pairs \cite{N2N} and methods that only require training sets consisting of single unpaired noisy images \cite{N2V}. 

However, the most flexible methods of them all are so-called blind zero-shot denoisers, which do not require a training set at all as they train only on the very image they are trying to denoise. This results in a plug-and-play denoiser that does not require a representative training set to be assembled, and as such is not biased by how well the dataset represents the target image. Another major advantage they offer is that they require limited technical skill to use, since they do not require the user to separately train a neural network on their own data.

State-of-the-art blind zero-shot denoisers can be broadly divided into two categories. The first are blind-spot networks where only a subset of the inputs participate in forward inference, and which typically get asymptotically more accurate over time. The other includes methods where all inputs participate in forward inference. This type of method typically results in a quick plateau in accuracy, followed by a precipitous drop as overfitting sets in and the model starts to learn the noise. 

Methods of the second type work in a variety of different ways and can be considered a category of exclusion. For example, Deep Image Prior (DIP) \cite{DIP} attempts to exploit the innate structure of a convolutional neural network to recover the image from random inputs. Another example is Noise2Fast (N2F) \cite{N2F} which attempts to learn to denoise the image from its downsamplings. In general, these denoisers learn to denoise the image quickly, but they require convergence criteria to avoid overfitting.

Blind-spot networks on the other hand, essentially treat denoising as an image inpainting task, where a pixel is inferred from its surroundings, excluding the pixel itself from consideration. Accordingly, since the neural network is blind to the pixels it is denoising, these methods are impeded from overfitting the data. Blind-spot methods, in particular Self2Self (S2S) \cite{S2S}, offer unparalleled accuracy although they can be quite slow. They also throw away the most valuable piece of information, the pixel itself, when determining the true signal at that location.

Here, we present a method that avoids this limitation and combines the strengths of blind zero-shot and blind-spot network denoisers. Our approach achieves higher accuracy and higher speed than S2S, the current gold standard. While retaining the structure of a blind-spot network, our method can see some of the input pixels during training and inference. Since our approach is not completely blind to the inputs, it is susceptible to learning the identity mapping, resulting in an overfitting like methods of the second type. Therefore, our method requires convergence criteria to be halted before accuracy plummets. To resolve this we introduce a domino tiling based pixel sampling approach we refer to as Pixel Domino Tiling, which we describe in the next section. Validated this way, our semi-blind spot network is an average three times faster and 0.28 PSNR more accurate than S2S, the current gold standard for blind zero-shot denoising. Further, we show that Pixel Domino Tiling can also be used to improve the accuracy of an existing method, Noise2Fast. This demonstrates the broader applicability of Pixel Domino Tiling.

\begin{figure*}
  \centering
  \includegraphics[width=0.9\linewidth]{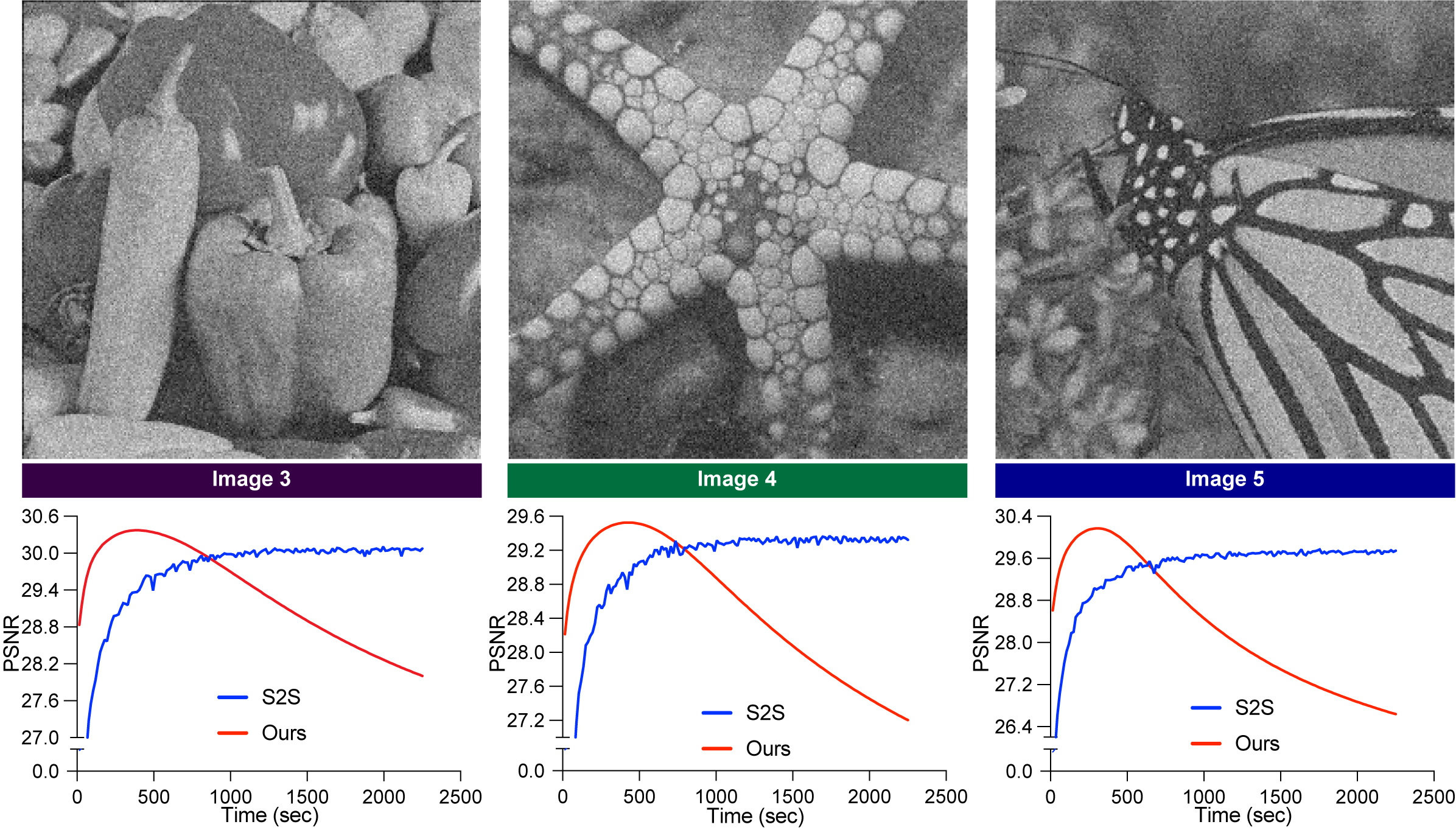}
\caption{\textbf{The accuracy progression of our technique.} PSNR over time of our modified blind-spot network vs. Self2Self (S2S) on image 3,4 and 5 from Set12 with $\sigma = 25$ Gaussian noise. Our method achieves an early peak that exceeds Self2Self's final accuracy before rapidly decreasing.}
\label{fig_sim}
  \label{fig:1}
\end{figure*}

\section{Theory}
Let $\mathbf{x} \in \mathbb{R}^{m \times n}$ be a noisy 2D image. Then we can write
\begin{align} \mathbf{x} = \mathbf{s} + \mathbf{n}. \end{align}
where $\mathbf{s}$ denotes the signal and $\bf{n}$ denotes the noise. Let $D \subseteq \mathbb{R}^{m \times n}$ be a domain of 2D images that we are interested in (e.g. images of mouse skeletal muscle). A denoiser on $D$ is a function $f:\mathbb{R}^{m \times n} \rightarrow \mathbb{R}^{m \times n}$ such that for $\mathbf{x} \in D$, we have that
\begin{align} f(\mathbf{x}) \approx \mathbf{s}. \end{align}

We can train a neural network to learn the function $f$. The simplest way to do this is using a representative training set of noisy/clean image pairs and training a CNN to map noisy images to their clean counterparts \cite{DnCNN}. Surprisingly, this also works for noisy/noisy image pairs, and with some mild assumptions if we train a neural network to map different noisy realizations of the same image to each other
\begin{align} \label{N2N} f(\mathbf{s} + \mathbf{n}_1) \rightarrow \mathbf{s} + \mathbf{n}_2, \end{align} 
it will also learn to denoise \cite{N2N}. We can also take it one step further and train an accurate denoiser on sets of unpaired noisy images \cite{N2V, N2S}. 

However, all these methods require the user to assemble some representative training set $K \subseteq D$. Not only is this cumbersome and requires technical skill, but it can introduce bias and lead to an inconsistent user experience depending on how well $K$ represents $D$. Even once trained, it cannot be expected to perform well on images outside $D$ without re-training.

For these reasons interest has grown in zero-shot methods, which denoise single noisy images without reference to any outside training set. More specifically, they train solely on the image they are attempting to denoise. Hence, training and inference are coupled and applied together on a per image basis. That is
\begin{align} K = D = \{ \mathbf{x} \}. \end{align}

Zero-shot denoisers are flexible and easy to use. Since they train from scratch on each image, $D$ is only limited by the set of images the neural network actually works on, rather than being constrained by the contents of some training set. If the method is also blind, in that it doesn't require a user supplied estimate of the noise distribution, this set grows even larger. The most accurate blind zero-shot denoisers are blind-spot networks, particularly S2S \cite{S2S}.

Blind-spot networks essentially treat denoising as an image inpainting problem, where at each iteration the neural network is blind to some subset of pixels  $P \subseteq \mathbb{N}_{\leq m} \times \mathbb{N}_{\leq n}$. The network is then trained to restore those pixels from the remaining information $\widetilde{P}$,
\begin{align} f \big(\left.\mathbf{x}\right|_{\widetilde{P}}  \big) \rightarrow \ \left. \mathbf{x} \right |_{P} . \end{align}

Excluding the pixels it is trying to restore prevents the network from learning the identity. This exclusion can be achieved in several ways. The current most accurate method, S2S \cite{S2S}, uses partial convolutions, in the true spirit of image inpainting. 

While blind-spot networks are the most accurate denoisers currently available, they tend to be slower than other methods since not all neurons participate in forward inference. Also, by excluding the pixel itself from its own prediction, they rob themselves of the most valuable predictor of the signal at that location.

An alternate method that avoids these limitations is DIP \cite{DIP}, which attempts to generate the image from randomly initialized weights. Since all neurons participate in forward inference, DIP is much faster than S2S. However, DIP quickly starts overfitting the data, and it is not trivial to establish convergence criteria. Another alternative is N2F \cite{N2F}, which maps downsamplings of the image to one another. N2F is by far the fastest blind zero-shot denoiser, however, accuracy pales in comparison to S2S.

N2F works by dividing the input image into two smaller images by a process known as checkerboard downsampling. More specifically, given an image $\mathbf{x}(i,j)$ indexed by its pixel co-ordinates, the checkerboard downsamplings $\mathbf{x}_\text{even}(i,j)$ and $\mathbf{x}_\text{odd}(i,j)$ are defined as

\begin{align} \label{CHECK} \mathbf{x}_\text{even}(i,j) &= \mathbf{x}(i,2j+(i \;\mathrm{mod}\; 2)), \\
\mathbf{x}_\text{odd}(i,j) &= \mathbf{x}(i,2j+(i \;\mathrm{mod}\; 2)+1). \end{align}
Noise2Fast then attempts to learn the mapping
\begin{align}\label{N2F0} f(\mathbf{x}_\text{even}) \rightarrow \mathbf{x}_\text{odd},  \\ f(\mathbf{x}_\text{odd}) \rightarrow \mathbf{x}_\text{even}. \end{align}
In essence, an input pixel gets mapped to one of its neighbours. The underlying idea is that the signal at a given pixel will in general be similar to its neighbours, this is based on a concept from an ealrier method, Neighbor2Neighbor \cite{Ne2Ne}.

So esentially we have two kinds of blind zero shot denosier, Blind-spot methods like S2S \cite{S2S}, which are slow, accurate and reliably converge, but don't use all available information. And methods like DIP and N2F, which are fast, use all available information, but require convergence criteria to be halted before overfitting sets in.We propose a middle ground between these two solutions. 

As stated previously, blind-spot networks train themselves to restore pixels from their surroundings. Since they are blind to the inputs they are trying to restore, they are prevented from learning the identity function.  However, this comes at the cost of discarding what should be the best predictor of the signal at that location during inference, the pixel itself.
\begin{figure*}[!t]
\centering
\includegraphics[width=0.9\linewidth]{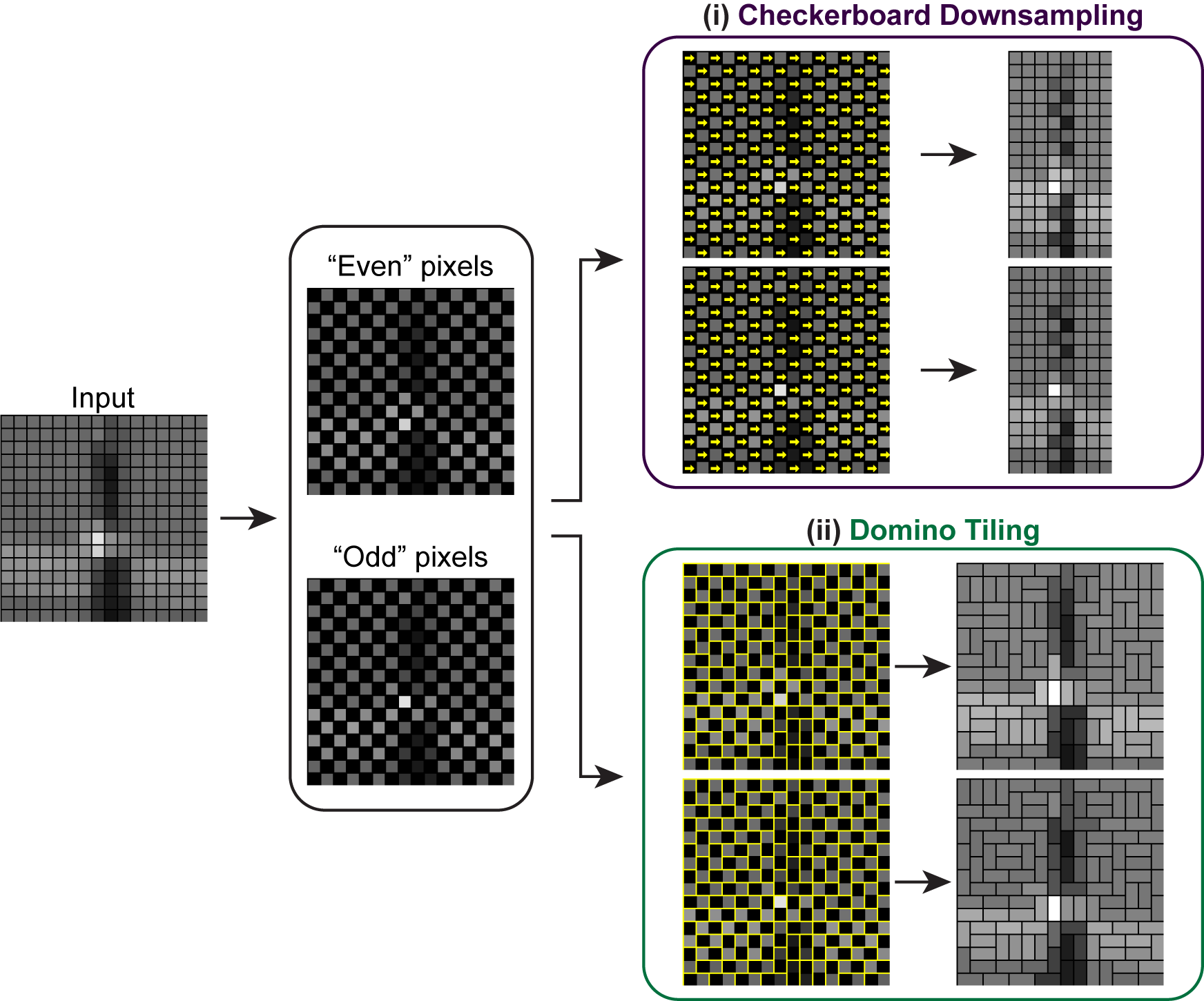}
\caption{\textbf{Pixel Domino Tiling contrasted with Checkerboard Downsampling.} The original input image is initially split into even and odd pixels for both procedures. In Checkerboard Downsampling \textbf{(i)}, the images are squeezed in to fill these gaps resulting in an image that is downsampled along one axis by a factor of two. With our Domino Tiling-based strategy \textbf{(ii)}, the pixel gaps are instead filled with neighbouring pixels (selected by solving the Linear Sum Assignment Problem described in Section 2). This preserves the original size proportions of the image and avoids distorting the pixel lattice relationship as much as Checkerboard Downsampling.}
\label{fig_sim}
\end{figure*}
A previous effort to resolve this issue is Blind2Unblind \cite{B2B}. While not a zero-shot technique, Blind2Unblind \cite{B2B} employs a trained mask mapper coupled with a creative loss function to re-incorporate missing information. However, to avoid learning the identity mapping, this method is still ultimately blind to the precise pixel values at the time gradients are updated.

So we set out to build a blind-spot network that could fully incorporate the actual pixel values. More specificlaly, we modified the traditional blind-spot network by adding a leaky mask, which randomly unmasks 0.1 \% of the pixels at each iteration. This has the effect of allowing the neural network to very rarely see the pixels it is trying to restore. In other words, we train our neural network to learn the mapping

\begin{align} f \big(\left.\mathbf{x}\right|_{\widetilde{P}}  \big) \rightarrow \ \left. \mathbf{x} \right |_{P \cup P_0}. \end{align}

Where $P_0 \subseteq \widetilde{P}$ samples 0.1 \% of the elements of $\widetilde{P}$. We then construct the output image using only those pixels that the neural network was able to see (i.e., from $\widetilde{P}$). Hence, our output denoised image exclusively contains pixels that were not masked during inference. See Fig. 1 for an overview of our modified blind-spot network, which we refer to as a semi-blind spot network.

Our observation is that under this scheme, images achieve a higher peak accuracy than S2S (Fig. 2), before overfitting sets in. 
Therefore, all we need is convergence criteria to realize these gains in accuracy.

We first investigated the checkerboard downsampling used by N2F, not as a tool for training, but as a validation scheme. To clairfy this distinction, recall that Noise2Fast adopts the following training/validation strategy

\begin{figure*}[!t]
\centering
\includegraphics[width=0.9\linewidth]{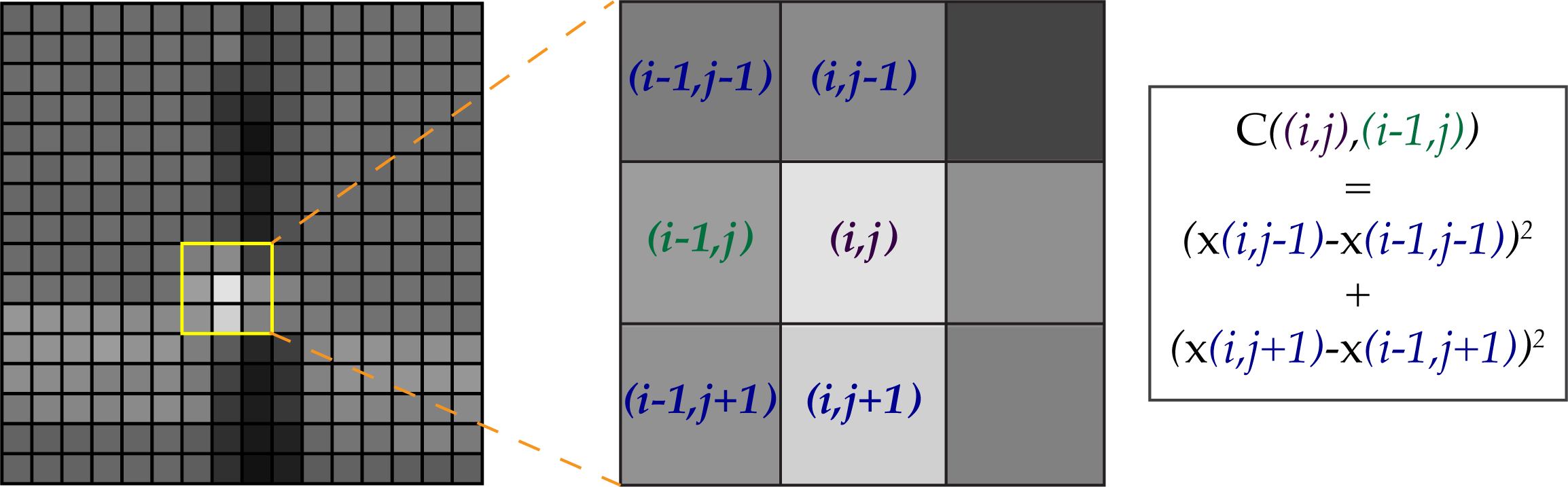}
\caption{\textbf{Pixel similarity cost function for Domino Tiling illustrated.} Utilizing nearby pixels as a proxy, our cost function calculates the local variance between the target pixel and each of the adjacent pixels. The idea is that the pixel with the lowest variance will be the most reasonable substitute.}
\label{fig_sim}
\end{figure*}

\begin{flalign}\label{N2F0} &\textbf{Train:} & f(\mathbf{x}_\text{even}) &\rightarrow  \mathbf{x}_\text{odd},  &\\ & \textbf{Validate:} & f(\mathbf{x}) & \rightarrow  \mathbf{x}. & \end{flalign}

This training strategy works because except at the most highly dynamic regions of the image, we have that

\begin{align} \mathbf{x}_\text{even} \approx \mathbf{x}_\text{odd}. \end{align}

Therefore, training a neural network to learn the mapping $f(\mathbf{x}_\text{even}) \rightarrow \mathbf{x}_\text{odd}$ teaches it to denoise in a similar fashion as Noise2Noise.

However, this validation strategy simply tests how well the neural network maps the noisy image back to itself, i.e., how well it matches the identity function. This is slightly counterintuitive; however, it works because, at a local level, the image itself looks so vastly different than its checkerboard downsamplings. Therefore, in N2F they can validate denoising efficiency on the original image as if it is unseen data. The point where the mapped image starts to diverge away from the original, is closely followed by the point where the neural network begins to overfit \cite{N2F}.

We observe however, that since this training/validation strategy essentially works on the idea that the downsamplings of the image do not locally resemble the image itself, it should be possible to reverse this scheme

\begin{flalign}\label{N2F0} & \textbf{Train:} & f(\mathbf{x}) & \rightarrow  \mathbf{x}, &\\ &\textbf{Validate:} & f(\mathbf{x}_\text{even}) &\rightarrow  \mathbf{x}_\text{odd},  & \end{flalign}
so that it trains on the full-sized image and validates on the downsamplings. However, since training a neural network to learn the identity function isn't particularly useful, we introduce our semi blind-spot approach to get
\begin{flalign}\label{DT} & \textbf{Train:} & f \big(\left.\mathbf{x}\right|_{\widetilde{P}}  \big) & \rightarrow \ \left. \mathbf{x} \right |_{P \cup P_0} &\\ &\textbf{Validate:} & f(\mathbf{x}_\text{even}) &\rightarrow  \mathbf{x}_\text{odd}.  & \end{flalign}
Although we got decent results with this validation scheme, we were quite far from achieving peak accuracy. Therefore, we sought to improve checkerboard downsampling by improving what we believe is its major flaw, the downsampling itself. Although \cite{N2F,5459271,7299621} note that there is significant across-scale self-similarity in natural images, the fact remains that downsampling fundamentally distorts the pixel lattice relationship. Therefore, when a neural network trained on full-sized images is applied to downsampled images (or vice-versa), the network will overestimate (or underestimate) the significance of neighboring pixels.

So we sought a way to achieve the benefits of checkerboard downsampling, without actually downsampling the image. Note that checkerboard downsampling consists of two steps, first you remove half the pixels in a checkerboard pattern, then you squeeze the remaining pixels in to fill in the gaps left by the removed pixels (see Fig. 3). We would like to avoid this second step, by instead filling the missing pixels in with some sensible value.

\begin{figure*}[!t]
\centering
\includegraphics[width=0.9\linewidth]{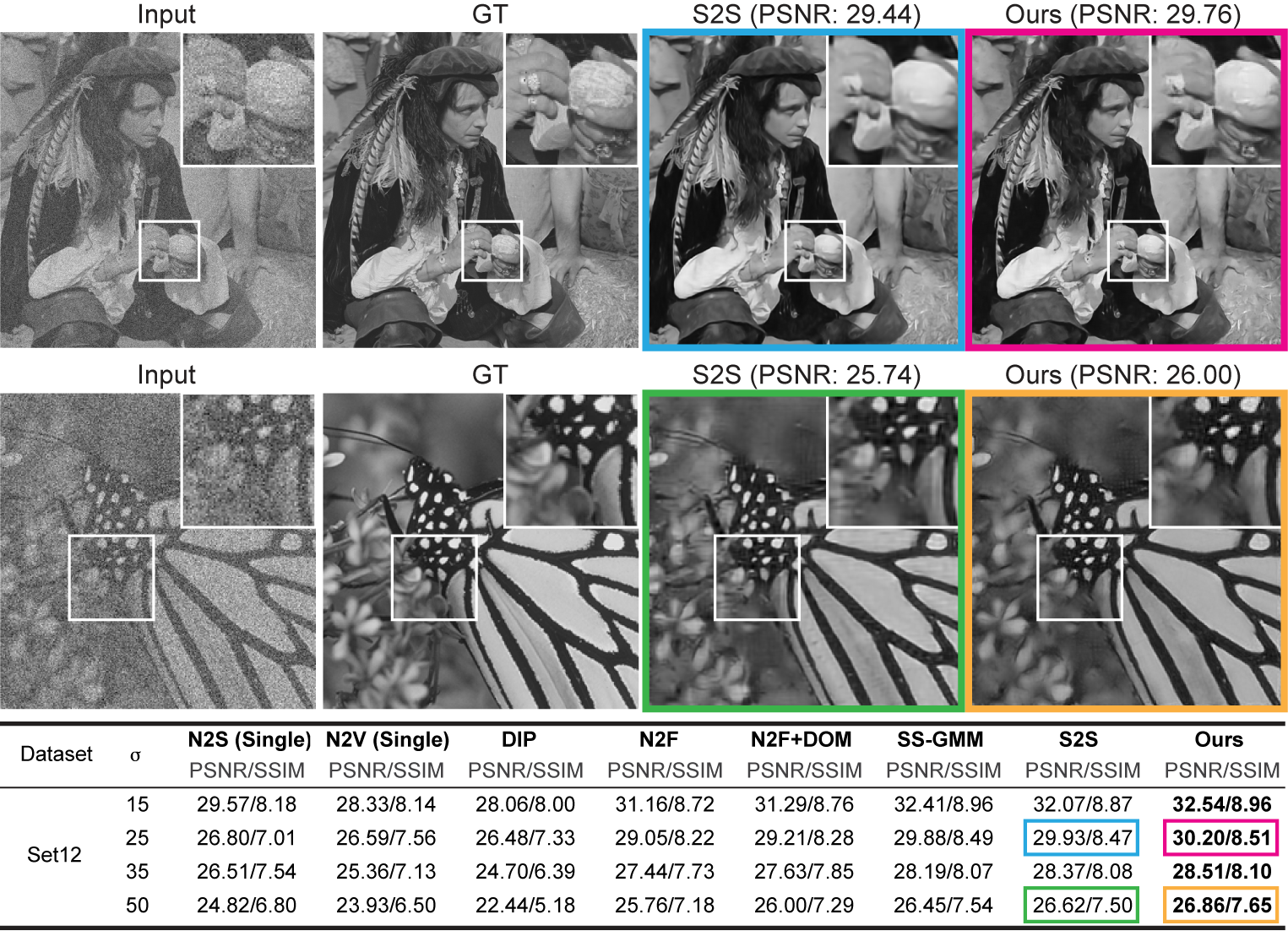}
\caption{\textbf{We compare our approach with other cutting-edge methods.} We compare the PSNR and SSIM of our method and other state-of-the-art blind zero-shot denoisers on synthetic Gaussian noise that is added to Set12 images. In the top row, we visually compare our method to Self2Self (S2S) on a Set12 image that has had $\sigma=25$ Gaussian noise added, and in the middle row, we compare on $\sigma=50$. In the bottom row, we show benchmarking results for the entire Set12 using each method.}
\label{fig_sim}
\end{figure*}

The easiest way to fill them in is to just replace each missing pixels with an average of its four immediate neighbors. Ultimately, this approach leads to unsatisfactory results (see ablation study Avg Nbr). This could be because the averaging process also denoises the pixels to some extent, so our image now appears less noisy and more blurry than the data our network was originally trained on.

To avoid this problem, we can fill the pixels in with a random neighboring value. Then each pixel will contain the same amount of noise as the original image. This leads to better results (see ablation study - Rand Nbr). However in highly dynamic regions, a randomly assigned neighbor is often very different from the pixel itself. So when a pixel is near a sharp edge in the image, we would like it to ``prefer'' to be filled in with a pixel that is more similar to itself to avoid going off the edge. Therefore we attempted a more deterministic approach where we try to swap in the neighbor that is in the direction of lowest variance (see ablation study - Best Nbr), by using the following cost function:

 \begin{align} \label{CF} &C\big((i,j),(i+c_1,j+c_2)\big) =  \\ 
 &  \quad \  \big( \mathbf{x}(i+c_2,j+c_1)-\mathbf{x}(i+c_1+c_2,j+c_1+c_2) \big) ^2 \\ &+ \big( \mathbf{x}(i-c_2,j-c_1)-\mathbf{x}(i+c_1-c_2,j-c_1+c_2) \big) ^2. \end{align}
Where $(c_1,c_2) \in \{(0,1),(0,-1),(1,0),(-1,0)\}$.

This is easier to understand visually (see Fig. 4), but essentially we are testing the average variance in each of the four cardinal directions, higher cost meaning that two pixels are less likely to share the same underlying signal.

Interestingly, calculating the domino tiling in this way reduced our accuracy as compared to our random approach. We believe that the reason this occurs is because the output image contains a significant amount of repeating pixels, particularly in edge-like regions where the direction of lowest variance traces around the edge contour. Ultimately this means that some parts of the image have access to a greater diversity of information than other parts.  

To avoid this, given an $m \times n$ image, we would like to constrain how we assign neighboring pixels such that the following two properties hold
\begin{enumerate}
    \item Each pixel gets assigned to an empty space
    \item No two pixels get assigned to the same empty space
\end{enumerate}

More formally, we would like to find a bijection $f:D \rightarrow L$ where
\begin{align} \label{LD}
D = \{(i,j):i+j = 0 \;\mathrm{mod}\; 2 \}, \\
L = \{(i,j):i+j = 1 \;\mathrm{mod}\; 2 \},
\end{align}
and where $f(i,j)$ is adjacent to $(i,j)$ for every $(i,j) \in D$. This is equivalent to finding a domino tiling of an $m \times n$ grid. 

The domino tiling problem, also known as the Dimer problem, is concerned with covering and $m \times n$ grid (or more generally a graph) with $2 \times 1$ dominos, so that all squares are covered by exactly one domino, and every domino covers two squares (See Fig. 3) \cite{DOM1}. Domino tilings have applications in statistical mechanics \cite{DOM2}, particularly in the study of phase transitions. They can also be generalized to higher dimensions \cite{DOM3} and to different polyominos \cite{DOM4}.

Returning to the checkerboard analogy, any domino laid on a checkerboard must touch one dark square and one light square, hence any domino tiling implies a bijection between the set of dark squares and the set of light squares. But the set of dark squares and the set of light squares are just $L$ and $D$ in \cref{LD}. Moreover, since dominos can only cover adjacent squares, satisfaction of the adjacency criteria is also guaranteed. Hence these two problems are equivalent.

\begin{figure*}[!t]
\centering
\includegraphics[width=0.9\linewidth]{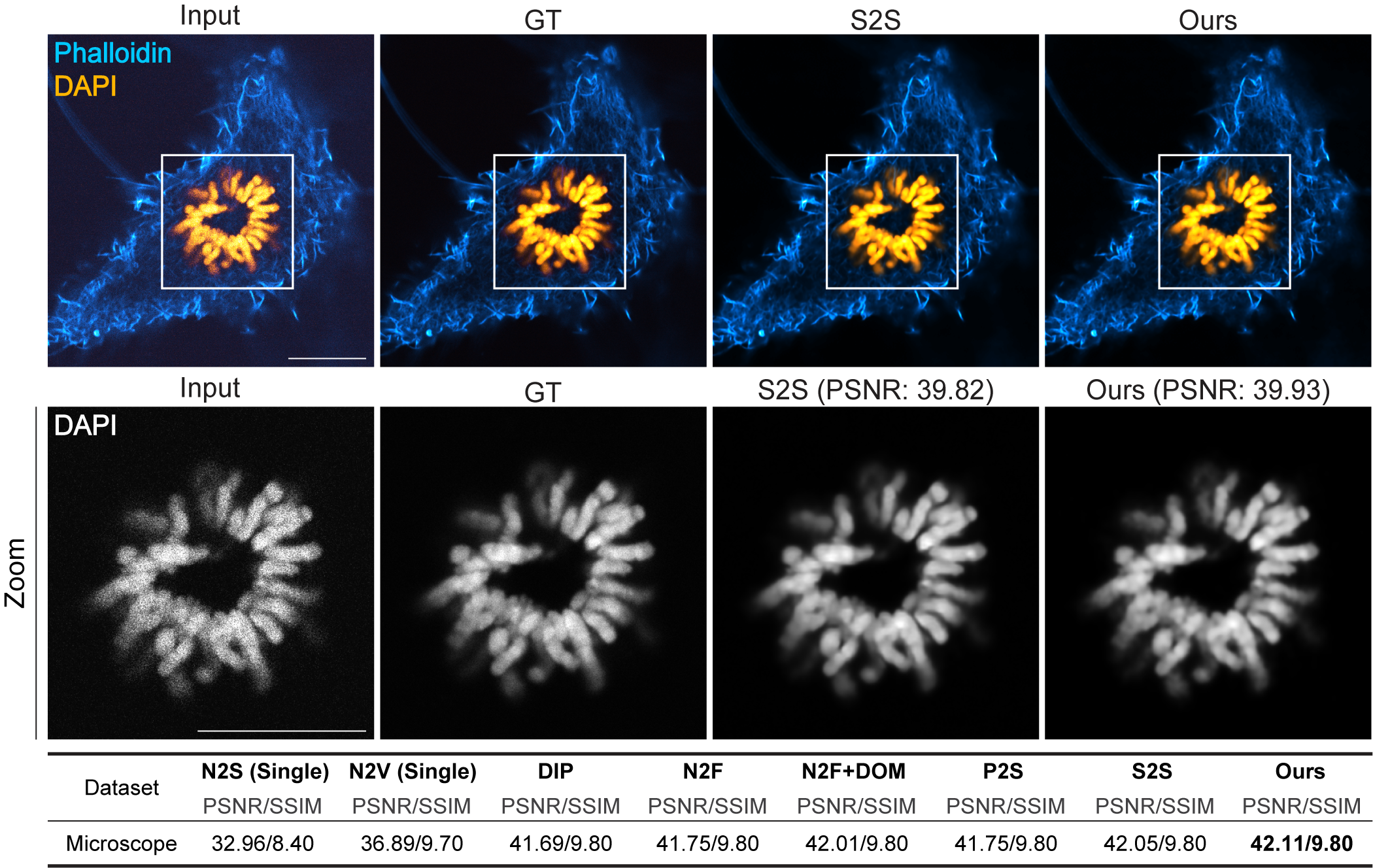}
\caption{\textbf{Denoising Comparison on real world microscopy images.} We compare the PSNR and SSIM of our method to other state-of-the-art blind zero-shot denoisers on real world Gaussian-Poisson distributed noise using images gathered from our laboratory. We stained RPE-1 cells with phalloidin-Alexa488 (Molecular Probes) and DAPI to label actin and DNA respectively. In the top row we visually compare our method to Self2Self (S2S) on one of these images, scale bars are 10 microns. In the bottom row we show benchmarking results for the entire set of images using each method.}
\label{fig_sim}
\end{figure*}

 This equivalence is useful to us, since Domino Tilings have already been thoroughly studied. Notice that no domino tiling can exist when there is an odd number of pixels. Therefore, we assume without loss of generality that there is an even number of pixels and we will pad our image as needed during training to ensure this. 
 
 For arbitrary graphs, just counting the number of domino tilings is $\# P$-complete, since it involves computing the permanent of the adjacency matrix. However, it is known from \cite{Kasteleyn} and \cite{Temperley} that this can be simplified for planar graphs to polynomial time, and for $m \times n$ rectangles the following formula exists
 
 \begin{align} \prod_{i=1}^{\lceil \frac{m}{2} \rceil }\prod_{j=1}^{\lceil \frac{n}{2} \rceil } \big( 4 \cos^2 \frac{\pi i}{m+1} + 4 \cos^2 \frac{\pi j}{n+1} \big). \end{align}
 
 Although we are less interested in counting domino tilings, than we are in generating them. We can of course easily generate domino tilings based on simple repeating patterns, however we are more interested in generating arbitrary domino tilings. If our desire is to uniformly sample from the set of all possible domino tilings, this can be achieved using Markov chains \cite{RandomDT2, RandomDT1}. However as previously stated, we would like to have some influence over the domino tiling we generate. 
 
To do this, we exploit a known equivalence and solve the domino tiling as a balanced linear assignment problem. A balanced linear assignment problem consists of $n$ agents and $n$ tasks, where each task must be assigned to one agent. No two agents can have the same task, and each assignment has a cost associated to it. The goal is to assign agents to tasks with a view to minimizing the cost.

If we take our dark squares to be agents, and our light squares to be tasks, and we use the cost function in \cref{CF}, then solving this linear assignment problem will find us a domino tiling that minimizes our cost function. There are many algorithms for solving linear sum assignment problems including the Auction Algorithm and the Hungarian Algorithm with runs in $\mathcal{O} (n^4)$ time. We opted for the more efficient Jonker-Volgenant Algorithm \cite{Jonker} which runs in $\mathcal{O} (n^3)$. 

Although solving this linear assignment problem is not instantaneous (it takes about 3 seconds for a 512x512 image), it still requires less than 1\% of the computation time required to denoise an image using our proposed method.

We call these solutions Pixel Domino Tilings of the image. We denote them ``even'' and ``odd'' depending on which pixels we built the domino tiling out of, similar to the convention for checkerboard downsampling \cref{CHECK}. If we validate our semi blind-spot network by checking how well our neural network maps ``odd'' Pixel Domino Tilings to their even counterparts (or vice versa) analagous to \cref{DT}, we can achieve a significant speed and accuracy advantage over S2S (see Results). We feel that our main contributions are as follows
\begin{figure*}[!t]
\centering
\includegraphics[width=0.9\linewidth]{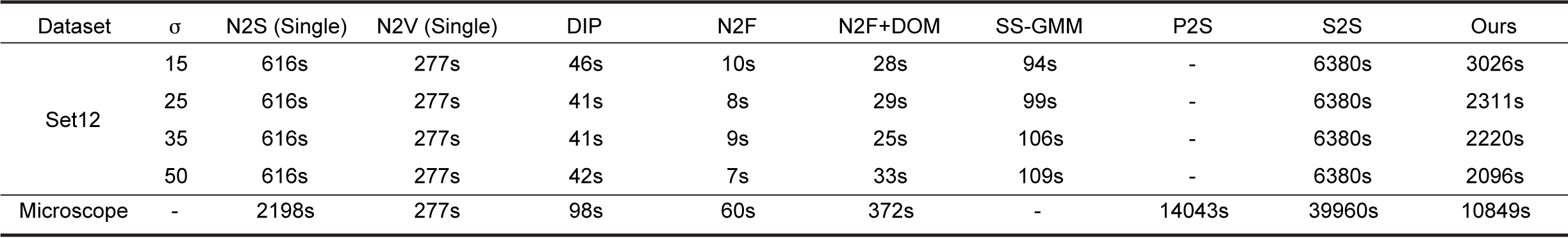}
\caption{\textbf{Speed comparison.} Per image time (in seconds) required to denoise each dataset with each compared method. Because they are specifically tailored to a certain type of noise, P2S was not benchmarked on Gaussian noise and SS-GMM was not benchmarked on Poisson noise.}
\label{fig_sim}
\end{figure*}

\begin{itemize}
\item{{\bf A novel semi-blind training architecture.}} We introduce the semi blind-spot network. By allowing our network to see a small percentage of the unmasked data at each iteration, the network can factor the pixel itself into its own prediction. By constructing the output solely from these unmasked pixels, we ensure that this additional information gets incorporated into the denoised image. This boosts accuracy at the cost of necessitating convergence criteria. The latter is addressed with our next contribution.
\item{{\bf A novel validation scheme based on Domino Tilings.}} We introduce Pixel Domino Tilings and use them to validate the convergence of our semi blind-spot network. Pixel Domino Tilings, improve upon checkerboard downsampling by allowing the neural network to train at-scale with the image. This approach not only gives us good convergence criteria for our presented network, but also can improve upon the accuracy of N2F itself at the expense of speed.
\item{{\bf Significant accuracy and speed gains over the current gold standard.}} We improve upon the accuracy of S2S by an average $ 0.28$ PSNR on synthetic Gaussian noise. We also realize an average three-fold gain in speed over standard S2S. This allows for unprecedented accuracy in a blind zero-shot denoiser, with a lower computational footprint.

\end{itemize}

\section{Related Work}

\subsection{Training Set Based Methods} The most accurate current denoisers are trained on a representative dataset consisting of noisy images and their clean counterparts. The earliest method to do this with a neural network was \cite{firstDn}. This was heavily refined in both the works of Mao et al. \cite{OtherDn} and Zhang et al. \cite{DnCNN} to achieve results that are still among the best achievable today. A more recent approach is \cite{GMS} which uses a two-module approach to push accuracy even further.

When clean ground truth images aren't available, Noise2Noise \cite{N2N} can be trained on paired noisy/noisy image data. A much faster version is presented in \cite{optica}. 

When more flexibility is desired, single-shot methods can be considered which allow the neural network to train on unpaired training sets of noisy images. An early method for doing this is N2V \cite{N2V} by using a so-called blind-spot network, this idea was built upon in \cite{N2S}, \cite{Laine}, \cite{BPAIDE}, and \cite{AP-BSN}. \cite{B2B} uses a mask mapper in conjunction with a novel loss function to unmask pixels in their blind-spot network.

Alternatives to blind-spot networks include \cite{Ne2Ne}, which trains a neural network to map pixels to their neighbors and Recorrupted-to-Recorrupted \cite{recor}, which attempts to generate noisy image pairs out of a single image. In \cite{unfold1} they present an interesting approach based on unfolding an image into its overlapping patches and associating them, although there is no publicly released code. In \cite{SVID} they attempt to build a method for creating noisy image pairs by subtracting the noisy image from a denoised pseudo ground-truth, randomly multiplying each element by $ \pm 1$ and then adding it back to the image. There are also adversarial GAN-like approaches such as in \cite{NIRN} and \cite{ISCL}. And in \cite{DIV} they present a VAE based approach.

\subsection{Zero-shot methods} 

One of the first zero-shot methods is BM3D \cite{4271520}. BM3D collects overlapping patches, clusters them and then finds a lower dimensional representation of those clustered patches. BM3D assumes Gaussian noise and requires a user supplied estimate of the standard deviation, therefore it is not blind. A more recent approach to non-blind denoising is \cite{LRRJS}, which uses a novel modelling strategy to better remove noise between similar patches. 
\begin{figure*}[!t]
\centering
\includegraphics[width=0.9\linewidth]{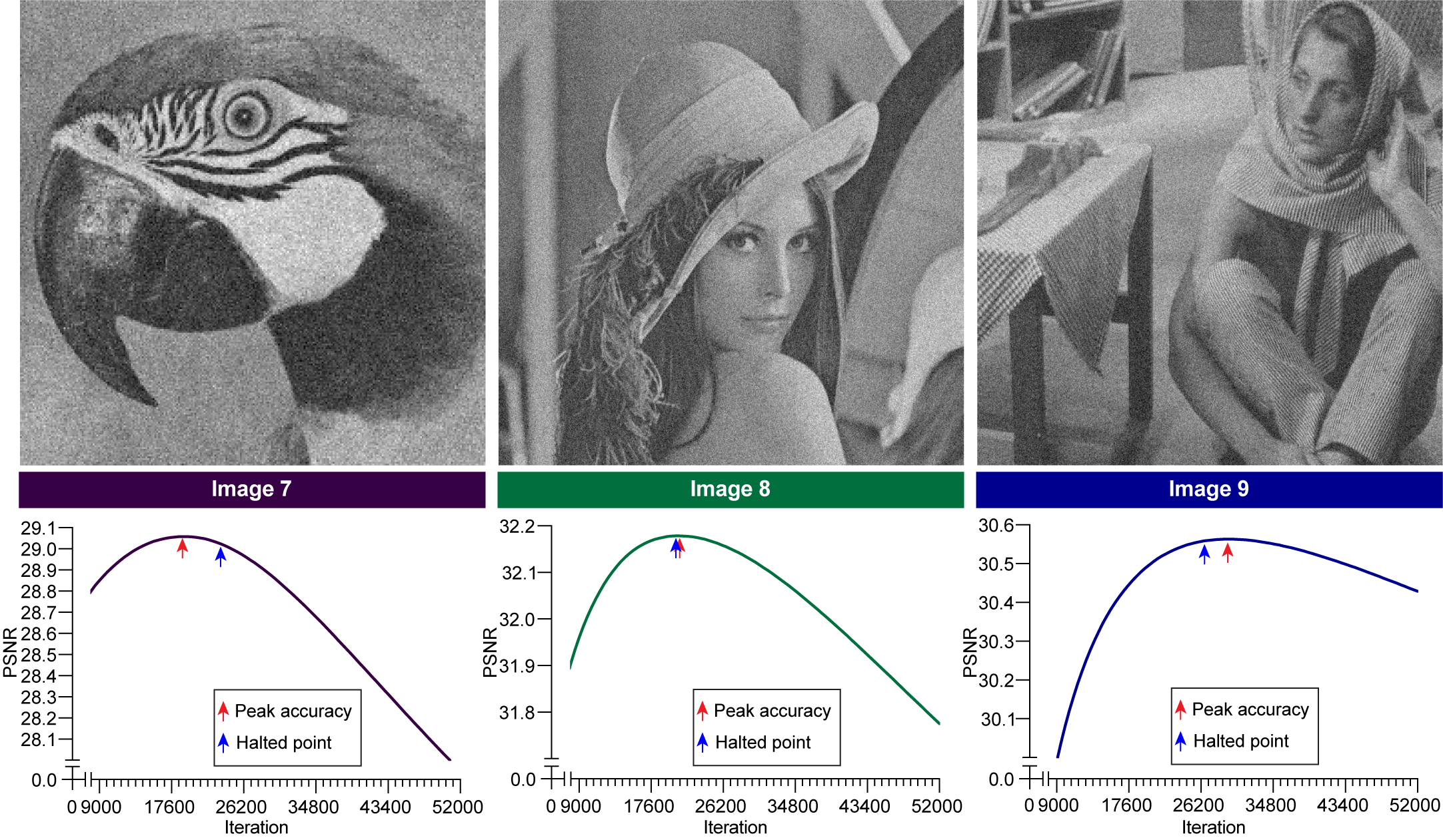}
\caption{\textbf{Effectiveness of our Pixel Domino Tiling validation scheme.} We compare the accuracy over time using our semi blind-spot network on three images from Set12 (Image 7, 8 and 9), using known ground truth images. Blue arrow indicates where our algorithm halted and red arrow indicates where theoretical maximum accuracy was achieved. All images were corrupted with $\sigma=25$ Gaussian noise.}
\label{fig_sim}
\end{figure*}
\subsection{Blind zero-shot methods} 
Noise2Self (N2S)\cite{N2S} is a blind-spot network and is the first such method to present a blind zero-shot version of itself. This is achieved by restricting the training set to a single image. Other methods can be similarly adapted, such as N2V \cite{N2V}. However, since these methods were originally tailored to train on large representative training sets, accuracy can be underwhelming and inconsistent with such adaptations.

The first method built and designed for blind zero-shot denoising is S2S \cite{S2S}. It is a blind-spot network that uses partial convolutions \cite{PConv} to mask out pixels, which is computationally expensive and slow.

DIP \cite{DIP} is the earliest non blind-spot network based approach. DIP trains a neural network to generate the image from randomly initialised weights. DIP initially denoises the image, however overfitting sets in quickly if it is run for too long.

In \cite{IDEA}, the authors present a method that selects attentional regions to denoise. It compares favourably to other methods within these attentional regions that the program selects. However, it does not denoise images as a whole.

Some recent methods are tailored specifically to low noise such as \cite{GMM} and \cite{ZSB}. \cite{GMM} assumes Gaussian distributed noise and therefore cannot be expected to perform well on microscopy images. On the other end of the spectrum, in \cite{poisson} they present a method specifically tailored to only work on Poisson distributed noise.

\section{Experiments}

\subsection{Compared Methods}
For Gaussian noise we compare our method against six other blind zero-shot denoisers: Noise2Self (N2S)\cite{N2S}, Noise2Void (N2V)\cite{N2V}, Self2Self \cite{S2S}, Deep Image Prior (DIP)\cite{DIP}, Noise2Fast (N2F) \cite{N2F} and SS-GMM \cite{GMM}. For Poisson noise, we replace SS-GMM (which is designed to work specifically on Gaussian noise) with Poisson2Sparse (P2S)\cite{poisson} (which is designed to work specifically on Poisson noise). We also add our own seventh method to each comparison, N2F+Domino (N2F+DOM), which is N2F where we replaced checkerboard downsampling with our domino tiling based approach. This serves to illustrate the broader applicability of domino tiling to generate noisy image pairs. 

We tested each method adhering as closely as possible to officially released software from the authors. We have also attempted to standardize image normalization between the methods since different methods tackle this in different ways. We now describe how each method was calibrated, including our own.

\subsubsection*{Self2Self} We use their officially published GitHub code written in tensorflow. We use their default settings of 150000 iterations and a learning rate of 1e-4. For the microscopy images we found that the method sometimes failed and the output was featureless, therefore we reduced learning rate  to 3e-5 for this dataset.

\subsubsection*{Noise2Self} We make use of their single shot denoising notebook on GitHub as-is, only adjusting the image normalization step for comparability with other methods.

\subsubsection*{Noise2Void} For Noise2Void used their ImageJ plugin. We calibrate it with a neighbourhood radius of 5, 64x64 patch size. We the run it with a batch size of 16, 10 steps per epoch and 10 epochs.

\subsubsection*{Deep Image Prior} Deep Image Prior is only a blind zero-shot method if we fix the maximum number of iterations. We set this number to 3000 to be consistent with the authors GitHub code.

\subsubsection*{Noise2Fast} We use the published GitHub code as-is, with no modificaton.

\subsubsection*{Noise2Fast+Domino}
For this method, we replaced the checkerboard downsamplings this method uses with pixel domino tilings. We took the four-image training set of checkerboard downsampled images and replaced it with a two image training set consisting of even and odd domino tilings of the input image. For better results we also adjusted the number of iterations per validation check from $100$ to $250$.

\subsubsection*{SS-GMM} We use their published matlab code as-is, only adjusting the image normalization step for comparability with other methods.

\subsubsection*{Poisson2Sparse} We use their published code as-is. We note that this method has a large memory footprint - to work on a $1024 \times 1024$ image we needed to use a GPU with 48GB of memory (RTXA6000). Our results may not be reproducible on a GPU with less memory.

\subsubsection*{Our method}

Our neural network is straightforward, we perform twelve $3x3$ partial convolutions each followed by ReLU activation. We then close with a 1x1 convolution plus sigmoid activation (Fig. 1). 

Before training starts, we compute the even and odd domino tilings for our image, that we will be using for validation. This consists of solving two linear sum assignment problems, one for the even pixels and one for the odd pixels using the cost function illustrated in Fig. 4.

At each step we take our input image, randomly mask $20\%$ of pixels and then feed this into our neural network. We then compute binary cross-entropy (BCE) loss between the pixels that were masked and the original noisy image, randomly unmasking $0.1 \%$ of pixels as described in the previous section.

At each iteration, we aggregate all pixels that the network was not blind to into a list for each pixel. Every $500$ iterations we take the average of each pixel's list and deem this the current output.

For validation, at each iteration we also randomly put either our odd or Pixel Domino Tiling through the network and aggregate it every $500$ iterations similar to the output. At every $500$ iterations we compare how well the neural network mapped the even domino tiling to the odd domino tiling and vice versa.

To explain how we quantify this, we will abuse terminology and call every 500 iterations an epoch.  At each epoch, we take the average of all the validation outputs for every iteration in that epoch. We then compute the percentage of pixels whose mean-squared error has gone up since the last epoch. We take a rolling average of the percentage calculated in 7 previous and 7 subsequent epochs. Once this rolling average has not gone up for $30$ more consecutive epochs, we terminate the program and output the image from 30 iterations ago.

\subsection{Benchmarking}
All benchmarking was conducted on a single RTX A6000 GPU.

\subsubsection*{Synthetic Gaussian noise}

 For synthetic additive white Gaussian noise bechmarking we use the commonly encountered Set12, where a separate script adds between $15$ and $50$ standard deviations of additive white Gaussian noise. As can be seen in Fig. 5 our method is more accurate than all tested methods for this dataset, beating S2S by an average $0.28$ PSNR and is also three fold faster than S2S. 

\subsubsection*{Real world microscopy}

We also compare performance on real world $1024 \times 1024$ microscopy images containing Gaussian-Poisson distributed noise. To obtain these images, RPE-1 cells were fixed with 4\% paraformaldehyde at room temperature for 10 min. The cells were then blocked with a blocking buffer (5\% BSA and 0.5\% Triton X-100 in PBS) for 30 min. Cells were washed with PBS and subsequently incubated with phalloidin-Alexa488 (Molecular Probes) and DAPI in blocking solution for 1 hour. After a final wash with PBS, the coverslips were mounted on glass slides by inverting them onto mounting solution (ProLong Gold antifade; Molecular Probes). For the fixed imaging in Fig. 6, single Z slice of cells were imaged on the Nikon Ti2E/AIR-HD25 scanning confocal microscope using a 60×/1.4 NA oil-immersion Plan-Apochromat lambda objective. Image acquisition was carried out with the resonance scan head with a single scan representing our noisy input at 1024px by 1024px and a 16x averaged scan representing our simulated ground truth. All images are displayed with auto scaled LUTs.  As can be seen in Fig. 6 our method outperforms all others and is three fold faster than S2S (Fig. 7).

\begin{figure}[!t]
\centering
\includegraphics[width=0.9\linewidth]{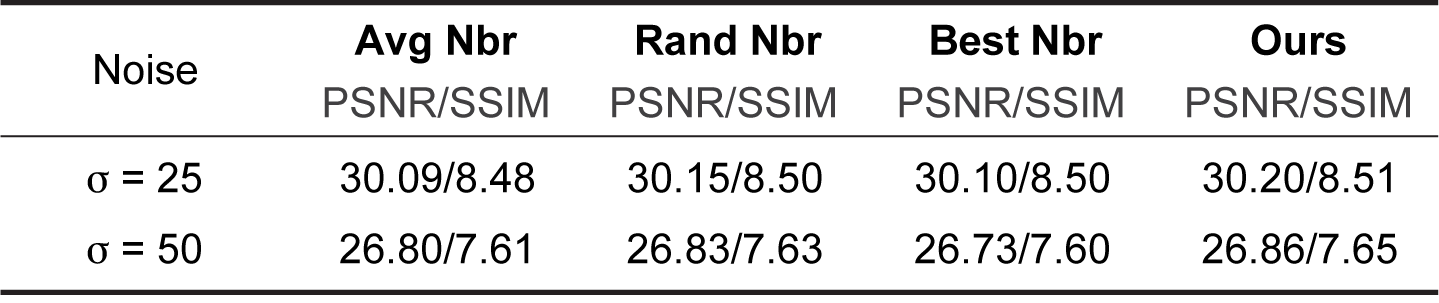}
\caption{\textbf{Ablation study on Set12.} We investigate alternative validation schemes that avoid the need to compute Pixel Domino Tilings, on Set12 with $\sigma=25$ and $\sigma=50$ Gaussian noise. In Avg Nbr we fill in pixel gaps by averaging all surrounding pixels. In Rand Nbr we randomly fill pixel gaps with one of its neighbors. In Best Nbr we select the lowest cost neighbor according to our cost function depicted in Fig. 4.}
\label{fig_sim}
\end{figure}

\subsubsection*{Domino Tiling}

We also investigate the utility of Pixel Domimo Tiling on its own. First, we integrate Pixel Domino Tiling into N2F, and show that in doing this we can improve its accuracy by an average $\approx 0.1$ PSNR on Gaussian noise and real world microscopy images (Figs. 5 and 6). Although this accuracy improvement comes at a significant speed cost, it establishes that the utility of Pixel Domino Tiling is not limited to the method we present in this study.

We also examine how good Pixel Domino Tiling based validation is at finding the peak accuracy in Fig. 8 and show that, although our validation strategy does not perfectly determine the peak accuracy, the difference between peak accuracy and the point where our validation strategy halts the algorithm is quite small.

\subsection{Ablation Study}
We test alternative ways to fill in the pixel gaps that avoid the need to use domino tilings. We first test whether we can just fill in pixel gaps by taking an average of its 4 immediate neighbours (Avg Nbr). We then try an approach where we select a neighbor at random to fill in the pixel gap (Rand Nbr). Finally, we test an approach where we always select the lowest cost neighbor according to our cost function in Fig. 3, without worrying about two pixels mapping to the same gap and hence avoiding the need for domino tiling (Best Nbr). As can be seen, in all cases we achieve higher accuracy using Pixel Domino Tiling (Fig. 9).

\section{Conclusion}
We present a blind zero-shot denoiser that outperforms the gold standard in terms of accuracy and speed. Our key innovation is realizing that we can achieve a higher accuracy by relaxing the ``blindness'' of a blind-spot network, and that we can resolve the overfitting issues this creates using a validation strategy based on computing Pixel Domino Tilings of the image. Our method achieves a noticeable improvement in PSNR across all tested datasets and noise levels. We achieve an average $0.28$ PSNR gain for synthetic Gaussian noise when compared to S2S the current gold standard. Although our gain is smaller for our microscopy dataset, it is worth noting that microscopy images often consist largely of featureless background on which significant accuracy gains are difficult to achieve.

 From a theoretical point of view, we believe that our Pixel Domino Tiling based validation strategy may be an asset to other denoising methods that require convergence criteria to avoid overfitting. Combined with our semi blind-spot network, we provide a computational framework that can potentially be applied to future denoising algorithms.

% Note that the IEEE does not put floats in the very first column
% - or typically anywhere on the first page for that matter. Also,
% in-text middle ("here") positioning is typically not used, but it
% is allowed and encouraged for Computer Society conferences (but
% not Computer Society journals). Most IEEE journals/conferences use
% top floats exclusively. 
% Note that, LaTeX2e, unlike IEEE journals/conferences, places
% footnotes above bottom floats. This can be corrected via the
% \fnbelowfloat command of the stfloats package.

% or


\ifCLASSOPTIONcompsoc
  \section*{Acknowledgments}
\else
  \section*{Acknowledgment}
\fi

We thank members of the Pelletier Lab for their scientific feedback during the project. We would like to thank Dr. Johnny Tkach and Thaisa Luup for carefully testing our code. JL was funded in part by the Ontario Graduate Scholarship (OGS) Program. AE was funded by a CIHR Foundation Post-Doctoral Fellowship (Funding Reference Number: 181763). The remainder of this work was funded by CIHR Foundation (FRN: 167279) and Krembil Foundation grants to LP which was used to fund JL, RP, and WH. LP is a Tier 1 Canada Research Chair in Centrosome Biogenesis and Function. The Network Biology Collaborative Centre at the LTRI is supported by the Canada Foundation for Innovation, the Ontario Government, and Genome Canada and Ontario Genomics (OGI-139). We would also like to thank Nikon for their support, LP's lab is a Nikon Centre of Excellence at the Lunenfeld-Tanenbaum Research Institute.

\section*{Data Availability}

Benchmarking datasets along with code and reproducibility instructions for the data in Fig. 5, 6 and 7 are available on our GitHub (https://github.com/pelletierlab/DominoDenoise). Note that all speed benchmarks were performed on an RTX A6000 GPU, and therefore results may vary according to GPU used. The scripts used to generate the data in Fig. 2, 8 and 9 are publicly available on our GitHub (DominoDenoise/AlternateScripts). Fig. 1, 3 and 4 are conceptual illustrations and do not make use of any datasets, however the image we use for illustrative purposes is a crop of an image available on our GitHub (DominoDenoise/Set12/12.tif).

\section*{Author Contributions Statement}
JL conceived of and coded the presented methods and wrote the manuscript. WH performed all benchmarking and prepared each of the figures. RP imaged the biological structures presented in Fig. 6 and helped prepare Fig. 6 and Fig. 5. AE participated in the developments that led to us using a LAP based approach to find domino tilings. LP encouraged JL to investigate denoising, supervised the findings of this work and funded the project. All authors assisted in writing the manuscript.

\ifCLASSOPTIONcaptionsoff
  \newpage
\fi



%\bibliographystyle{IEEEtran}
{\small
\bibliographystyle{naturemag}
\bibliography{egbib}
}


\begin{IEEEbiography}[{\includegraphics[width=1in,height=1.25in,clip,keepaspectratio]{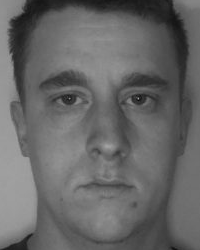}}]{Jason Lequyer}
Jason Lequyer received his MSc degree in Mathematics and is currently pursuing his PhD at the University of Toronto. He is a member of the Pelletier lab at the Lunenfeld-Tanenbaum research institute at Mount Sinai Hospital, Toronto. His research interests include self-supervised learning, computer vision and game theory. He is currently focused on building tools that facilitate the visual analysis of the myriad of biological materials processed by the Pelletier lab using its vast collection of powerful microscopes.
\end{IEEEbiography}

\begin{IEEEbiography}[{\includegraphics[width=1in,height=1.25in,clip,keepaspectratio]{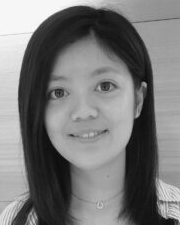}}]{Wen-Hsin Hsu}
Wen-Hsin Hsu received the MSc degree in Biomedical Sciences and the PhD degree in Tissue Engineering and Regenerative Medicine from National Chunghsing University, Taichung, Taiwan, in 2014 and 2018, respectively. She is a postdoctoral fellow with the Lunenfeld-Tanenbaum Research Institute at Toronto, Ontario, Canada. Her research involves studying the genetic underpinnings of organelle positioning in cells, using cell biology techniques in conjunction with AI based computational imaging pipelines.
\end{IEEEbiography}

\begin{IEEEbiography}[{\includegraphics[width=1in,height=1.25in,clip,keepaspectratio]{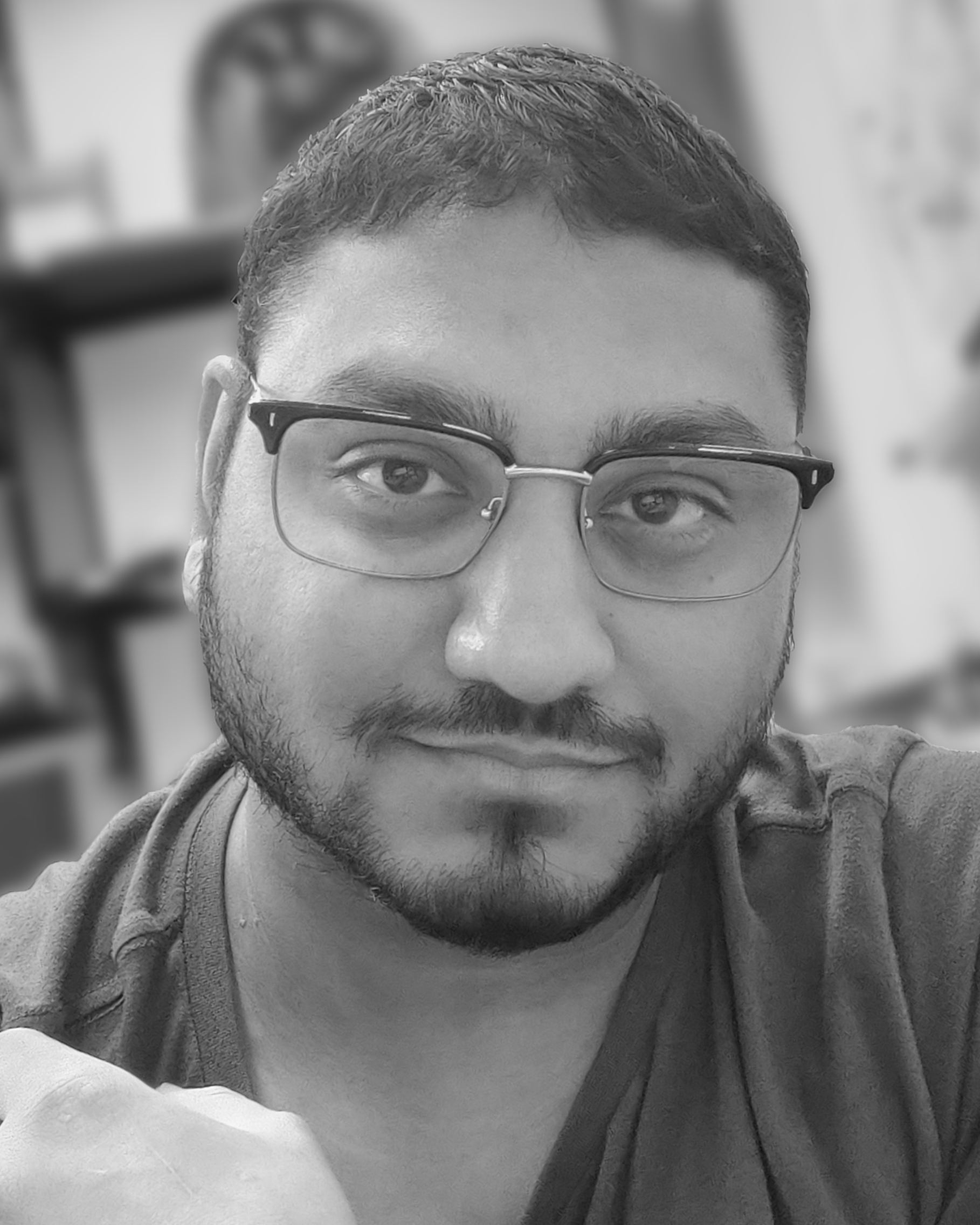}}]{Reuben Philip}
Reuben Philip received his BSc and MSc degree in cell and molecular biology from the University of Toronto Scarborough, Toronto, Canada. He is currently pursuing his PhD studies at the University of Toronto Department of Molecular Genetics, Toronto, Canada. His research interests include genome engineering and coupling automated microscopy pipelines with image analysis to uncover the pheno-genetic landscape of centrosome aberrations in cancer.
\end{IEEEbiography}

\begin{IEEEbiography}[{\includegraphics[width=1in,height=1.25in,clip,keepaspectratio]{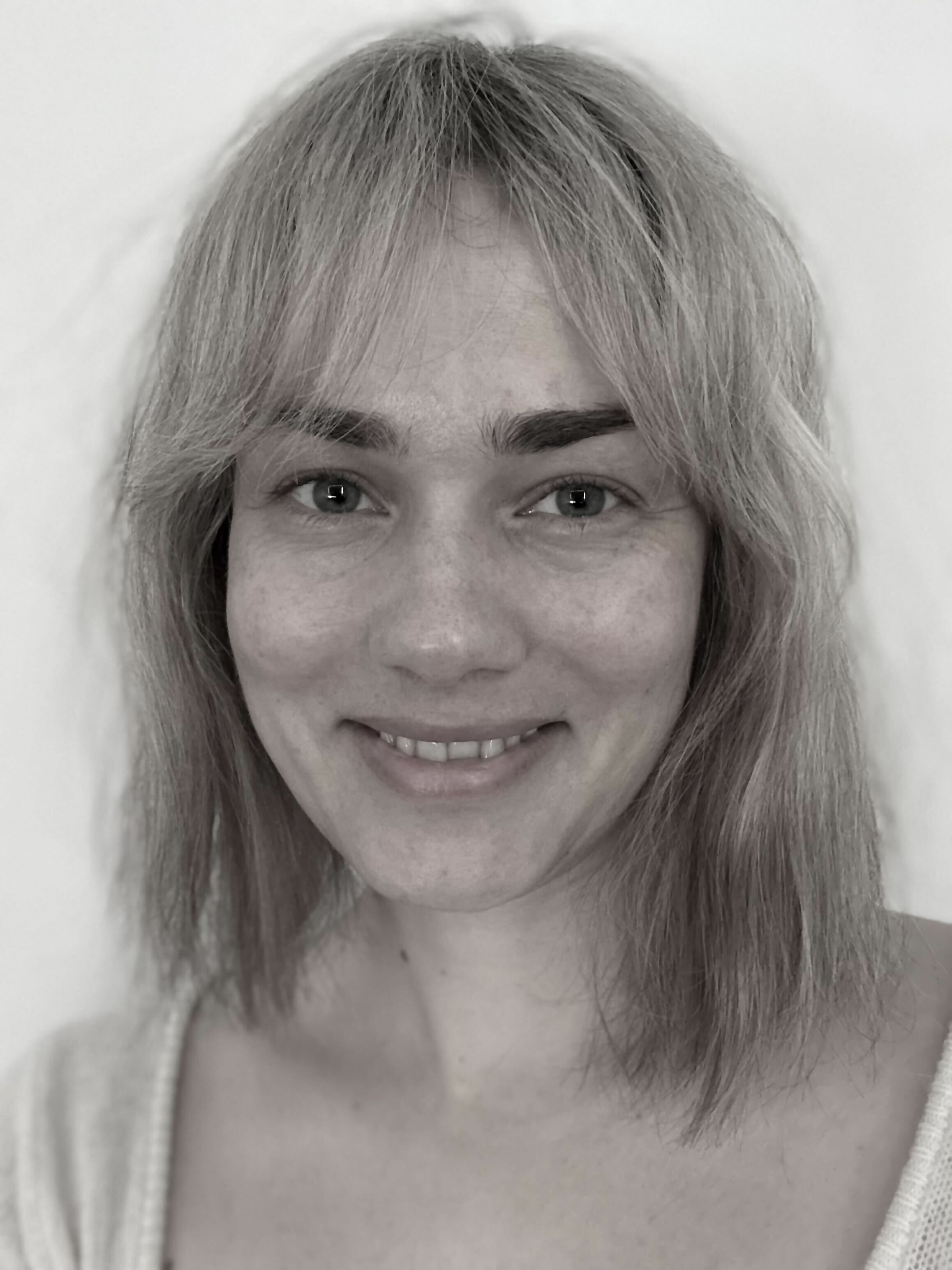}}]{Anna Christina Erpf} Anna Christina Erpf received her MSc and PhD in Cell and Developmental Biology from the Ludwig Maximilian University of Munich, Bavaria, Germany, in 2013 and 2020, respectively. She is currently a CIHR funded postdoctoral fellow (Funding Reference Number: 181763) at the Lunenfeld-Tanenbaum Research Institute, Toronto, Ontario, Canada. Her research interests include cell and developmental biology, high-throughput visual screening, microscopy, and image analysis.
\end{IEEEbiography}

\begin{IEEEbiography}[{\includegraphics[width=1in,height=1.25in,clip,keepaspectratio]{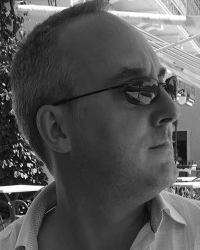}}]{Laurence Pelletier}
Dr. Laurence Pelletier obtained his PhD from Yale. He is a Senior Investigator at the Lunenfeld-Tanenbaum's Centre for Systems Biology, where he studies molecular mechanisms in cells that regulate centrosomes and cilia biogenesis and function in human cells. A better understanding of these fundamental cellular processes has important implications for cancer and myriad other diseases. He works with the most powerful light microscopes in the world, combined with state of the art computational imaging tools to help elucidate these processes.
\end{IEEEbiography}

%\newpage


%\vfill


\end{document}